\documentclass[pmlr,twocolumn,10pt]{jmlr} 

\usepackage{booktabs}
\usepackage{siunitx}

\usepackage[switch]{lineno}

\theorembodyfont{\upshape}
\theoremheaderfont{\scshape}
\theorempostheader{:}
\theoremsep{\newline}

\jmlrworkshop{Machine Learning for Health (ML4H) 2025} 

\usepackage[utf8]{inputenc} 
\usepackage[T1]{fontenc}    
\usepackage{microtype}      
\usepackage{nicefrac}       

\usepackage{enumitem}

\usepackage{titletoc}
\usepackage[page,header]{appendix}

\usepackage{url}            
\usepackage{hyperref}
\hypersetup{
    colorlinks,
    citecolor=blue,
    filecolor=blue,
    linkcolor=blue,
    urlcolor=blue,
}

\usepackage{graphicx}
\usepackage{chngcntr} 
\begingroup
	\expandafter\ifx\csname pdfsuppresswarningpagegroup\endcsname\relax\else\global\pdfsuppresswarningpagegroup=1\relax\fi
\endgroup

\usepackage{makecell}
\usepackage{wrapfig}
\usepackage{booktabs}
\usepackage{multirow}
\usepackage{colortbl}
\usepackage{tablefootnote}
\usepackage{array}
\newcolumntype{P}[1]{>{\centering\arraybackslash}p{#1}}
\usepackage{tikz}

\usepackage{amsmath,amssymb,amsfonts}
\usepackage{dsfont}
\usepackage{scalerel}
\usepackage{bm}
\DeclareMathOperator*{\argmin}{argmin} 
\DeclareMathOperator*{\mean}{mean}
\usepackage{cancel}

\usepackage{caption}
\usepackage{subcaption}
\usepackage{titlesec}
\titlespacing\section{0pt}{3pt plus 2pt minus 0pt}{3pt plus 2pt minus 0pt}
\titlespacing\subsection{0pt}{2pt plus 2pt minus 0pt}{2pt plus 2pt minus 0pt}

\newcommand{\operation}[1]{\operatorname{#1}}

\title[Synthetic Data Reveals Generalization Gaps in Correlated Multiple Instance Learning]{Synthetic Data Reveals Generalization Gaps\\in Correlated Multiple Instance Learning}

\author{%
    \Name{Ethan Harvey}$^1$ \Email{ethan.harvey@tufts.edu}\\
    \Name{Dennis Johan Loevlie}$^1$ \Email{dennis.loevlie@tufts.edu}\\
    \Name{Michael C. Hughes}$^1$ \Email{michael.hughes@tufts.edu}\\
    \addr $^1$Department of Computer Science, Tufts University, Medford, MA, USA
}

\begin{document}

\setlength{\abovedisplayskip}{2pt plus 3pt}
\setlength{\belowdisplayskip}{2pt plus 3pt}

\maketitle

\begin{abstract}
    Multiple instance learning (MIL) is often used in medical imaging to classify high-resolution 2D images by processing patches or 
    classify 3D volumes by processing slices.
    However, conventional MIL approaches treat instances separately, ignoring contextual relationships such as the appearance of nearby patches or slices that can be essential in real applications.
    We design a synthetic classification task where accounting for adjacent instance features is crucial for accurate prediction.
    We demonstrate the limitations of off-the-shelf MIL approaches by quantifying their performance compared to the optimal Bayes estimator for this task, which is available in closed-form.
    We empirically show that newer correlated MIL methods still do not achieve the best possible performance when trained with ten thousand training samples, each containing many instances.
\end{abstract}
\paragraph*{Data and Code Availability}
Synthetic data and Python code are available at \url{https://github.com/tufts-ml/correlated-mil} and have been integrated into \href{https://github.com/Franblueee/torchmil}{torchmil}~\citep{castro2025torchmil}.

\paragraph*{Institutional Review Board (IRB)}
Our study uses synthetic data and does not require IRB approval.

\section{Introduction}
\label{sec:introduction}

Many prediction tasks in medical imaging involve visual data with varying cardinality, resolution, or dimensionality.
For example, inputs may consist of high-resolution 2D images (e.g., histopathology images) or 3D image volumes (e.g., CT or MRI scans).
In these scenarios, a common approach is to divide each image into smaller 2D patches or slices known as \emph{instances}, obtain per-instance representations, and then aggregate scores or representations across instances to make one prediction for the whole image~\citep{ilse2018attention,han2020accurate,shao2021transmil,harvey2023probabilistic}.
Building predictors that aggregate one coherent prediction from many instance representations is known as 
\emph{multiple instance learning} (MIL)~\citep{quellec2017multiple,dietterich1997solving,maron1997framework}.
MIL offers a practical framework for handling weakly labeled data.

\begin{figure*}[t!]
  \centering
  \includegraphics[width=1\linewidth]{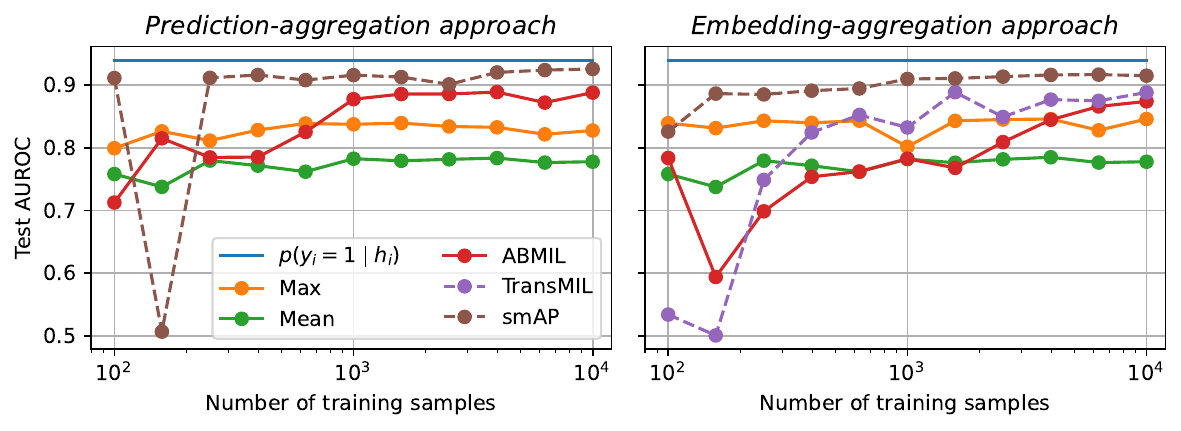}
  \caption{Test AUROC as a function of training set size $N$. All data is drawn from our Shifted Mean MIL data-generating process for binary classification, with $R{=}3, \Delta{=}2$.
  Conventional MIL approaches (Max, Mean, ABMIL) cannot match the Bayes estimator $p(y_i = 1 \mid h_i)$ as they do not account for dependencies between instances within a bag. 
  Surprisingly, even with $N{=}10000$, correlated MIL approaches (TransMIL~\citep{shao2021transmil}, smAP~\citep{castro2024sm}) do not reach the ceiling set by the Bayes estimator. smAP comes close, but bootstrapping reveals the Bayes estimator maintains a statistically significant advantage (mean of AUROC difference is 0.014, 95\% confidence interval of [0.007, 0.022] does \emph{not} include zero or any negative values).
  \textbf{Takeaway: Our work reveals a need for data-efficient MIL that better accounts for context between instances.}
  }
  \label{fig:varying_training_set_size}
\end{figure*}

Conventional MIL approaches broadly treat instances separately and independently.
This assumption ignores the spatial and contextual relationships between adjacent patches or slices. Accounting for these relationships can be critical for accurate prediction in medical applications.
To address this problem, recent work has proposed \emph{correlated MIL}~\citep{shao2021transmil} to model dependencies between instances. Others have built upon this direction~\citep{castro2024sm}. Assessing the capabilities and limits of such methods remains an open problem.

In this work, we take a synthetic data approach to better understand the importance of spatial and contextual relationships between adjacent instances in multiple instance learning.
Our contributions are:

\begin{itemize}[leftmargin=*,noitemsep,topsep=0pt]
    \item We design a novel synthetic dataset called \emph{Shifted Mean MIL} to represent key challenges in MIL for medical imaging: (1) only some features are discriminative, (2) only a few instances in each bag signal whether it should be positive class, and (3) context from nearby instances matters, as the information from an individual instance may be statistically ambiguous.
    \item We derive the optimal Bayes estimator for this dataset and use its predictions as a gold standard for comparing how well MIL methods perform.
    \item We demonstrate that even recent correlated MIL methods designed to account for context do not achieve the best possible performance on our toy task, as shown in  Fig.~\ref{fig:varying_training_set_size}.
\end{itemize}
These contributions suggest concrete opportunities for future work to improve correlated MIL, perhaps via improved inductive biases or regularization strategies.
\section{Related Work}
\label{sec:related_work}

\noindent\textbf{Multiple instance learning.} 
MIL is a branch of weakly supervised learning where a variable-sized set of instances has a single label.
Early MIL approaches used simple, non-trainable operations such as max or mean pooling to aggregate instance representations \citep{pinheiro2015image,zhu2017deep,feng2017deep}.
Recent work has proposed attention-based pooling \citep{ilse2018attention}.
Several works have extended attention-based pooling while maintaining permutation-invariance \citep{li2021dual,lu2021data,keshvarikhojasteh2024multi}.
Correlated MIL~\citep{shao2021transmil} extends traditional MIL by modeling relationships between instances within a bag, allowing the pooling operation to capture morphological and spatial information rather than treating instances as independent.
\citet{castro2024sm} proposed a smoothing operator to introduce local dependencies among neighbors.
\citet{shao2025multiple} showed that transfer learning with MIL approaches improves generalization.

Most similar in spirit to our work are the \emph{algorithmic unit tests} for MIL proposed by \citet{raff2023reproducibility}.
They suggest three synthetic classification tasks designed to reveal whether learned models violate key MIL assumptions, such as a bag is positive if and only if one or more instances have a positive label.
Our new data focuses on cross-instance dependency, which was not examined by \citet{raff2023reproducibility}.

\textbf{Adjacent context in deep learning.}
Several prior works have introduced architectural modifications to better capture dependencies between adjacent patches in 2D images or slices in 3D images. 
Shifted windows allow for cross-window connections in vision transformers (ViT) \citep{liu2021swin}.
Weight inflation transfers pre-trained weights from lower- to higher-dimensional model (e.g., 2D to 3D CNNs) \citep{carreira2017quo,zhang2022adapting}.

\begin{figure*}[t!]
    \centering
    \includegraphics[width=\linewidth]{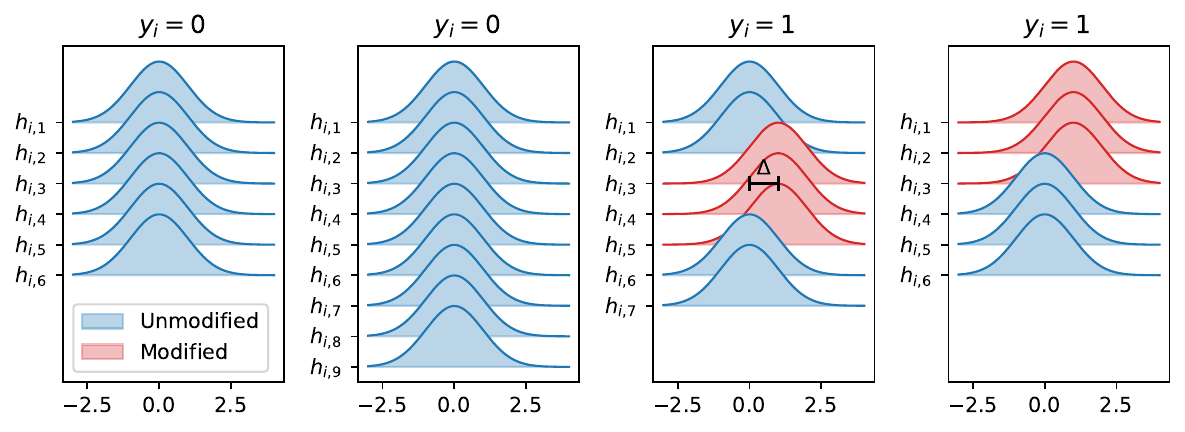}
    \caption{Example data-generating distributions for a discriminative feature for negative ($y_i{=}0$) and positive ($y_i{=}1)$ ``bags'' of $S_i$ instances drawn from our Shifted Mean MIL synthetic data.
    Setting $R{=}3$ means context around modified instances (in red) can help.
    In our experiments, we set $R{=}3$, $S_{\text{low}}{=}15$, $S_{\text{high}}{=}45$, $K{=}1$, $M{=}768$, $\mu{=}0$, and $\sigma{=}1$. We study how prediction quality changes as we vary training set size $N$ (Fig.~\ref{fig:varying_training_set_size}) and class separation $\Delta$ (Fig.~\ref{fig:varying_separation}).
    }
    \label{fig:toy_data_visualization}
\end{figure*}

\section{Background}
\label{sec:background}

\subsection{Multiple Instance Learning}

To train MIL models, the training dataset $\mathcal{D} = \{(x_{i,1:S_i}, y_i)\}_{i=1}^N$ consists of $N$ labeled bags. Each bag is a set of $S_i$ independent instance feature vectors $\{x_{i,1}, \ldots, x_{i,S_i}\}$ with a single label $y_i$.
Among deep MIL pipelines, there are two broad paradigms: \emph{embedding-aggregation} and \emph{prediction-aggregation}. Both approaches have neural architectures consisting of three parts: an encoder, a pooling operation, and classifier. They differ in the ordering of these components.

In the \emph{embedding-aggregation} approach, the order is encode, pool, then classify.
First, each instance's $B$-channel raw image $x_{i,j} \in \mathbb{R}^{B \times W \times H}$ is encoded into a representation vector $h_{i,j} = f(x_{i,j}) \in \mathbb{R}^M$.
Second, a pooling operation $\sigma$ (e.g., max, mean, or attention-based pooling) aggregates all $S_i$ instance representations $\{h_{i,1}, \ldots, h_{i,S_i}\}$ into a single representation vector $z_i = \sigma(h_{i,1:S_i}) \in \mathbb{R}^M$.
Usually, this pooling is \emph{permutation-invariant}.
Finally, the bag level representation vector $z_i$ is classified into a predicted probability vector over $C$ classes, $c(z_i) \in \Delta^C \subset \mathbb{R}^C$. We can denote the ultimate prediction as $\hat{y}_i = c(\sigma(f(x_{i,1:S_i}))$. In this notation, applying $f$ to a set yields another set containing a mapping of each instance.

In the \emph{prediction-aggregation} approach, the ordering of $c$ and $\sigma$ is swapped. A separate prediction score (e.g., a logit or probability vector) is produced for each of the $S_i$ instances separately, and then pooling determines the final prediction, $\hat{y}_i = \sigma(c(f(x_{i,1:S_i})))))$.

Ultimately, in either approach, model parameters for all three parts (encoder, pooling, and classifier) are trained to minimize binary or multi-class cross entropy averaged across all data: $\frac{1}{N} \sum_{i=1}^N \ell^{\text{CE}}(y_i, \hat{y}_i)$, where $\hat{y}_i$ is a function of input features and parameters.

\subsection{Pooling Methods}

The design of the pooling layer $\sigma$, which aggregates across instances, is generally most important for understanding how spatial context is incorporated. We describe several architectures below. We focus on \emph{embedding-aggregation} for concreteness; a translation to \emph{prediction-aggregation} is straightforward. Here, we take as input a set of embeddings $h_i := h_{i,1:S_i}$ for bag $i$. Each instance $j$ in the bag is embedded as vector $h_{ij} \in \mathbb{R}^M$.

\textbf{Max and Mean.} Two simple poolings find the maximum or mean \emph{\textbf{element-wise}} for $M$-dim.~vectors:
\begin{align}
    z_{i} = \max_{j=1,\dots,S_i} h_{ij}, \quad \text{or} \quad
    z_{i} = \mean_{j=1,\dots,S_i} h_{ij}.
\end{align}

\textbf{Attention-based pooling.}
Attention-based pooling~(ABMIL)~\citep{ilse2018attention} assigns an attention weight $a_{ij}$ to each instance, then forms bag-level embedding vector $z_i$ via a weighted average:
\begin{align}
    z_i = \sum_{j=1}^{S_i} a_{ij} h_{ij},~~a_{ij} = \frac{\exp\left(u^\top \tanh\left(U h_{ij}\right)\right)}{\sum_{k=1}^{S_i} \exp\left(u^\top \tanh\left(U h_{ik}\right)\right)},
    \notag 
\end{align}
where the weights $a_{ij}$ are non-negative and sum to one: $\sum_j a_{ij} = 1$; $a_{ij} {\ge} 0$ for all $j$. 
Here, vector $u \in \mathbb{R}^L$ and matrix $U \in \mathbb{R}^{L \times M}$ are trainable parameters.

\textbf{Smooth attention pooling.}
Smooth attention pooling (smAP) \citep{castro2024sm} uses a smoothing operation to add local interactions between instance embeddings. The smoothed embeddings $g_i \in \mathbb{R}^{S_i \times M}$ for all $S_i$ instances are obtained by solving an optimization problem
\begin{align}
\texttt{Sm}(h_i) &= \argmin_{g_i} \alpha\mathcal{E}_D(g_i) + (1-\alpha) \|h_i - g_i\|^2_F,
\end{align}
where $\alpha \in [0, 1)$ controls the amount of smoothness, $\| \cdot \|_F$ denotes the Frobenius norm, and
\begin{align}
    \mathcal{E}_D(g_i) &= \frac{1}{2} \sum_{j=1}^{S_i} \sum_{k=1}^{S_i} A_{ijk} \|g_{ij} - g_{ik}\|^2_2.
\end{align}
Here, $A_i \in \mathbb{R}^{S_i \times S_i}$ is an adjacency matrix defining local relationships between instances and $\| \cdot \|^2_2$ denotes the squared Euclidean norm aka ``sum of squares''.

\textbf{Correlated MIL pooling.}
\citet{shao2021transmil}'s 
transformer-based correlated MIL (TransMIL) allows instance interactions to inform pooling.
First, TransMIL uses convolutions over instances in a pyramidal position encoding generator to model dependencies. Second, interactions between \emph{all pairs} of instances are captured via multi-head self-attention. For layer $\ell$ and head $h$, there's a
$S_i {+} 1 \times S_i {+} 1$ attention matrix, where rows sum to one and weight $j,k$ is:
\begin{align}
    a_{i,j,k}^{(\ell,h)} \propto \exp \left( \left(q_{i,j}^{(\ell,h)}\right)^\top k_{i,k}^{(\ell,h)} / \sqrt{d} \right).
\end{align}
Here, each instance $j$ has $L$-dim.~embeddings for query $q_{i,j}^{(\ell,h)} {=} W_Q^{(\ell,h)} h_{i,j}^{\ell-1}$, key $k_{i,j}^{(\ell,h)} {=} W_K^{(\ell,h)} h_{i,j}^{\ell-1}$, and value $v_{i,j}^{(\ell,h)} {=} W_V^{(\ell,h)} h_{i,j}^{\ell-1}$.
Propagating embeddings via attention-weighted value averages over several layers and heads allows instance features to interact flexibly to inform the ultimate bag-level embedding.

\section{Shifted Mean MIL Synthetic Data}
\label{sec:methods}

We propose a new data-generating process designed to mimic several key challenges in real-world multiple-instance medical imaging tasks:
\begin{itemize}[leftmargin=*,noitemsep,topsep=0pt]
    \item Across the whole dataset, only a few features of many are discriminative ($K$ of $M$).
    \item For each positively-labeled bag, only a few instances are relevant ($R$ of $S_i$) and they are adjacent in a known 1D listing of all $S_i$ instances.
    \item Context matters. Adjacent instances together provide stronger statistical signal than any one relevant instance's discriminative feature value alone.
\end{itemize}

\noindent The generative process for bag $i$ first draws the bag's binary label and the number of instances in the bag
\begin{align}
    y_i &\sim \operation{Bern}(q_{+}), \quad 
        S_i \sim \operation{Unif}(\{S_{\text{low}}, \dots, S_{\text{high}}\}).
\end{align}
Next, for negative bags we sample all features $m$ for all instances $j$ independently from a common Gaussian:
\begin{align}
 h_{i,j,k} \mid  y_i {=} 0 \sim \mathcal{N}( \mu, \sigma^2).
\end{align}
For positive bags, most instances and features are sampled from this same Gaussian. However, for the $K$ discriminative features, we select $R$ adjacent instances (using $u_i$ to denote the starting index) and sample these from a Gaussian with \emph{shifted mean}:
\begin{align}
\label{eq:hu_given_y}
    u_i \mid y_i {=} 1 &\sim \operation{Unif}(\{1, \dots, S_i - R + 1\}),
    \\ \notag 
    h_{ijk} \mid u_i, y_i {=} 1
    &\sim 
        \begin{cases}
        \mathcal{N}(\mu + \Delta, \sigma^2), 
        & \hspace{-.2cm} \text{if $j \in [u_i, u_i {+} R {-} 1]$} \\
        & \text{~and $k$ is discrim.}
        \\
        \mathcal{N}(\mu, \sigma^2), 
        & \hspace{-.2cm}
        \text{otherwise}.
    \end{cases}
\end{align}
Here $\Delta{>}0$ indicates the magnitude of shift for discriminative features. Setting $R{>}1$ indicates that context helps.
Given a fixed $\mu$, bags drawn from this process are more challenging to classify (even with knowledge of the true process) when $\Delta$ is smaller, $R$ is smaller, $\frac{K}{M}$ is smaller, and $\sigma$ is larger.

This data-generating process is illustrated in Fig.~\ref{fig:toy_data_visualization}, depicting only one feature that is discriminative.
In each positive bag, a different contiguous block of $R{=3}$ instances draw from the shifted mean Gaussian.
If future work wanted to model correlations between features within an instance, the sampling of vector $h_{ij}$ in Eq.~\eqref{eq:hu_given_y} could be modified  to draw from a multivariate Gaussian with a non-diagonal covariance matrix.



\section{Bayes Estimator}
Given a data-generating process, a \emph{Bayes estimator} is a decision rule that minimizes the posterior expected loss with respect to the data-generating distribution \citep{degroot1970optimal,murphy2022BayesianDecisionTheory}.
It is an oracle upper bound on performance.
By comparing conventional or recent MIL methods to the Bayes estimator for our synthetic dataset, we can quantify how close they come to the best possible performance.

Given a new bag $h_i$ containing $S_i$ instances and assuming our data-generating process defined above, a Bayes estimator for class label probability is:
\begin{align}
    p(y_i=1 \mid h_i) = 
    \frac{p(h_i \mid y_i=1) p(y_i=1)}{p(h_i)}.
\end{align}
Each term on the right-hand side can be computed in closed-form. For brevity we omit how random variable $S_i$ cancels out here; see App.~\ref{sec:distribution_over_number_of_slices} for details.
The denominator term is given by the sum rule:
\begin{align}
\notag 
\resizebox{\linewidth}{!}{$
    \displaystyle
    p(h_i) = p(h_i \mid y_i{=}0)p(y_i{=}0) + p(h_i \mid y_i{=}1)p(y_i{=}1).
$}
\end{align}
where we recall $p(y_i{=}1)$ is $q_{+}$. 
The class-conditional likelihood for the negative class factors over instances:
\begin{align}
    \notag 
    p(h_i \mid y_i=0) = 
    \textstyle \prod_{j=1}^{S_i} \prod_{k=1}^{M} \text{NormPDF}(h_{ijk} \mid \mu, \sigma^2).
\end{align}
Each positive bag has a latent segment of $R$ consecutive relevant instances. The class-conditional likelihood marginalizes out the unknown index $u$:
\begin{align}
\notag 
\resizebox{\linewidth}{!}{$
    \displaystyle
    p(h_i \mid y_i{=}1) = \sum_{u=1}^{S_i-R+1} \left[ 
    p(u \mid y_i{=}1)
    \prod_{j=1}^{S_i} \prod_{k=1}^{M} p(h_{ijk} \mid u, y_i{=}1) \right] .
$}
\end{align}
The two lines of Eq.~\eqref{eq:hu_given_y} provide the necessary PDF values to evaluate the right hand side.
\section{Experimental Results}
\label{sec:experimental_results}

\textbf{Setup.}
In our experiments, we sample bag labels uniformly ($q_+{=}0.5$) and the number of instances per bag uniformly between $S_{\text{low}}{=}15$ and $S_{\text{high}}{=}45$.
We use $M{=}768$ features to match the size of ViT-B/16 embeddings~\citep{dosovitskiy2020image} and use only a single discriminative feature ($K{=}1$).
We set $R{=}3$ so context matters, and fix $\mu{=}0, \sigma{=}1$.
For main results in Fig.~\ref{fig:varying_training_set_size}, we fix $\Delta{=}2$.
For results varying $\Delta$, see the Appendix.

For each train set size $N$, we draw a training dataset of $N$ bags from our data-generating process. We further draw a separate dataset of $\frac{1}{4}N$ bags for a validation set used to select hyperparameters.
After selecting a final model at each $N$,
we report AUROC performance on a common test set of 1000 bags.



For each MIL method, we try both \emph{prediction-aggregation} and \emph{embedding-aggregation} approaches when possible.
We train models for 1000 epochs, with potential for early stopping.
We explore a range of possible hyperparameters for each method, including learning rate in $\{10^{-1}, 10^{-2}, 10^{-3}, 10^{-4}\}$ and weight decay in $\{10^{0}, 10^{-1}, 10^{-2}, 10^{-3}, 10^{-4}, 10^{-5}, 10^{-6}, 0\}$.
Both early stopping and hyperparameter selection seek to maximize validation set AUROC.

Examining the results of our synthetic dataset experiments in Fig.~\ref{fig:varying_training_set_size}, key findings are:

\begin{itemize}[leftmargin=*]
    \item \textbf{Conventional MIL cannot match the Bayes estimator when context matters} ($R{=}3$).
    Even given $N{=}10000$ bags, conventional deep MIL approaches (Max, Mean, ABMIL~\citep{ilse2018attention})  deliver test AUROC at least 0.04 below the Bayes estimator in Fig.~\ref{fig:varying_training_set_size}. Trends over $N$ do not suggest this gap will close with more data.
    
    \item \textbf{TransMIL cannot match the Bayes estimator when $R{=}3$.}
    This is surprisingly, since TransMIL purports to handle context between instances. TransMIL's AUROC is at least 0.03 below the Bayes estimator at any tested $N$ value.
    
    \item \textbf{smAP cannot quite match the Bayes estimator when $R{=}3$, though it is consistently the best method at almost all $N$.}
    At almost all tested $N$ above 500, smAP~\citep{castro2024sm} scored within 0.02 AUROC of the Bayes estimator.
    At the largest training set size of $N{=}10000$ in Fig.~\ref{fig:varying_training_set_size}, the \emph{prediction-aggregation} variant of smAP reached its highest test set AUROC.
    To determine whether the AUROC difference from our Bayes estimator to this best smAP model was statistically significant, we performed bootstrap analysis (see App.~\ref{sec:bootstrap_analysis}). This analysis revealed that the 95\% confidence interval of the AUROC difference did not include zero or any negative values, so we conclude that smAP remains inferior to the Bayes estimator even at $N{=}10000$.
    We expect this performance gap to increase for $R{>}3$ because the chain graph adjacency matrix used in smAP limits instance interaction to immediate neighbors.

    \item \textbf{The ability to capture adjacent instance context is a primary reason for the gap.}
    To verify that the context-dependent $R{=}3$ setting is what is important, we verified that for data drawn from a context-free $R{=}1$ version of our process, many MIL methods including max pooling, ABMIL, and TransMIL can all match the Bayes estimator with handcrafted parameters (see App.~\ref{app:handcrafted_parameters}).

    \item \textbf{The key limitation appears to be in the training process, though architecture changes may help too.}
    We were able to \emph{handcraft} neural network parameters for TransMIL that closely match the Bayes estimator when $R{=}3$, as in Fig.~\ref{fig:handcrafted_parameters3} in the Appendix.
    However, the training of pooling and classification layers yields consistently suboptimal predictions, even when $N{=}10000$. We conjecture that better regularization beyond simple weight decay would help. Some architecture changes are likely also beneficial if they introduce useful inductive bias.   
    To demonstrate this latter point, in App.~\ref{app:handcrafted_parameters}, we show how adding context to ABMIL via convolutions over instances can help when $R{=}3$.    
\end{itemize}

\section{Conclusion}
\label{sec:conclusion}

We designed an easy-to-implement synthetic dataset designed to mimic key challenges in MIL for medical imaging, especially the need for context from nearby instances.
We then demonstrated the limited ability of conventional MIL on such data, by quantifying the performance gap between the optimal Bayes estimator.
Correlated MIL methods like TransMIL are still notably worse than optimal even with a 
labeled dataset of 10000 bags.
Very recent methods like smAP come closer, but still fall short.



\textbf{Outlook.} Our work reveals a key gap that must be overcome for MIL to succeed on real medical data. 
Many popular 3D brain scan datasets where MIL could help contain labeled datasets far smaller than the largest $N$ tested here. For example, RSNA~\citep{flanders2020construction}, as used in~\citet{castro2024sm}, has only 1150 scans for training and evaluation.
We hope our synthetic dataset enables the development of correlated MIL methods that can be trained effectively with limited labeled data and make a difference in disease detection and treatment.
\newpage

\subsection*{Acknowledgments}

The authors gratefully acknowledge support from the Alzheimer’s Drug Discovery
Foundation and the U.S. National Institutes of Health (grant \# 5R01NS134859-02). MCH is also supported
in part by the U.S. National Science Foundation (NSF) via grant IIS \# 2338962. 
\bibliography{main}

\newpage
\appendix
\onecolumn

\counterwithin{table}{section}
\setcounter{table}{0}
\counterwithin{figure}{section}
\setcounter{figure}{0}

\makeatletter 
\let\c@table\c@figure
\let\c@lstlisting\c@figure
\let\c@algorithm\c@figure
\makeatother

\section{Experiments Varying the Class Separation \texorpdfstring{$\Delta$}{Delta}}
\label{app:varying_separation}

Here, we study how the prediction quality of various MIL methods varies with the class separation parameter $\Delta$, which controls how similar the discriminative feature values are in the data-generating process for the positive class (mean $\mu + \Delta$, variance $\sigma^2$) and negative class (mean $\mu$, variance $\sigma^2$).

Experiments here use $R=3$, $N=400$, and $\sigma=1$. 
Given these settings, the chance of a positive discriminative feature value landing closer to the negative mean ($\mu$) than the positive mean ($\mu+\Delta$) is $Pr( h_{ijm} \leq \mu + \frac{1}{2} \Delta ) = \Phi( \frac{(\mu + \frac{1}{2} \Delta) - (\mu + \Delta)}{\sigma} ) = \Phi( -\frac{1}{2}\Delta)$, where $\Phi$ is the standard Normal CDF. When $\Delta=2$ as in the main paper, this is 0.1587; when $\Delta=3$, this is 0.0068; when $\Delta = 4$, this is 0.0228; when $\Delta = 5$, this is 0.0062. For $\Delta \geq 4$, any individual discriminative feature has over 97\% chance to correctly indicate the class label without much need for its surrounding context.




Indeed, in the actual experimental results below, as $\Delta$ increases, we see all methods improve and approach near perfect classification, except for the Mean pooling baseline.

However, for modest $\Delta$ values like 2 or below, there's a noticeable gap between the ideal Bayes estimator and the actual method performance. 
We expect the relatively small gap in AUROC between smAP at the largest $N$  and the Bayes estimator shown in Fig.~\ref{fig:varying_training_set_size} for $\Delta=2$ would notably increase as $\Delta$ gets smaller: note below a gap of roughly 0.1 between smAP and the Bayes estimator at $\Delta=1$.

\begin{figure}[htbp!]
  \centering
  \includegraphics[width=1\linewidth]{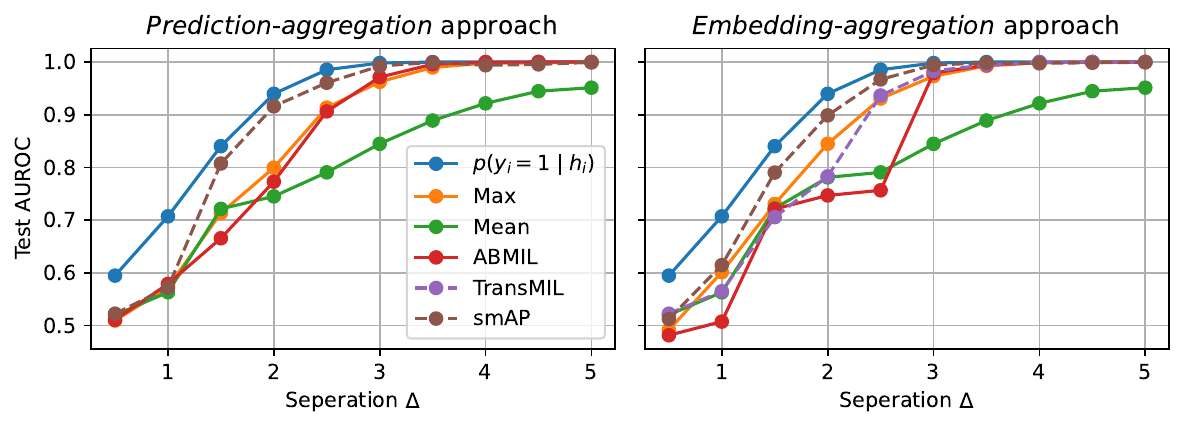}
  \caption{Test AUROC as a function of separation $\Delta$. All data drawn from our Shifted Mean MIL data-generating process fro binary classification, with $R{=}3$ and $N{=}400$.
  }
  \label{fig:varying_separation}
\end{figure}

\section{Handcrafted Parameters}
\label{app:handcrafted_parameters}

We set classifier weights corresponding to each discriminative feature to one, all other weights to zero, and the bias parameter to $-\frac{\Delta}{2}$. For attention pooling, we use the linear part of the tanh function to create attention weights proportional to each feature’s linear score. For instance convolutions, we set the center $R$ weights to one and all other weights and biases to zero to sum up each feature’s linear score over $R$ instances.

\begin{figure*}[htbp!]
  \centering
  \includegraphics[width=1\linewidth]{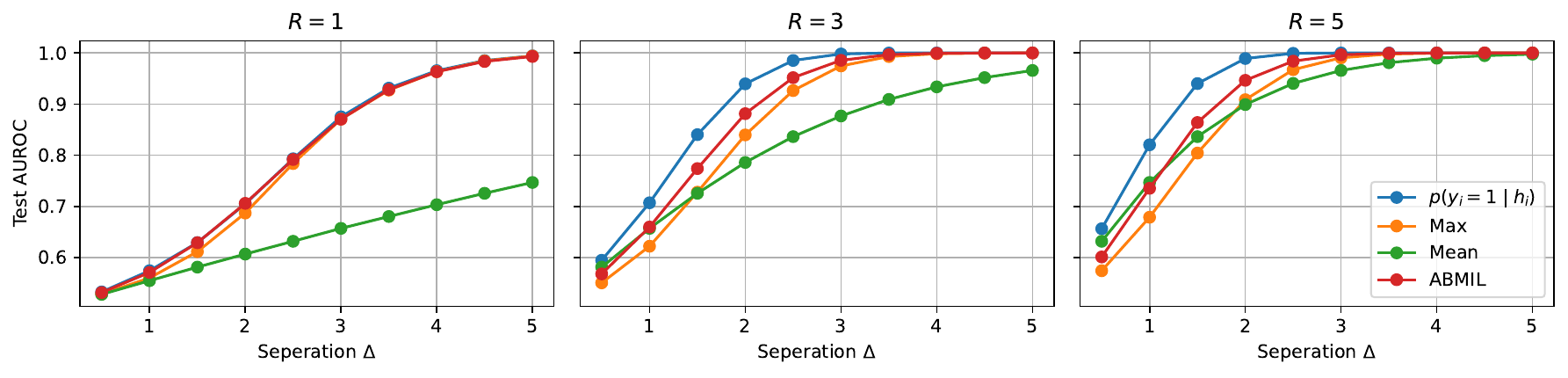}
  \caption{Handcrafted parameters (\emph{prediction-aggregation} approach).
  }
  \label{fig:handcrafted_parameters1}
\end{figure*}
\begin{figure*}[htbp!]
  \centering
  \includegraphics[width=1\linewidth]{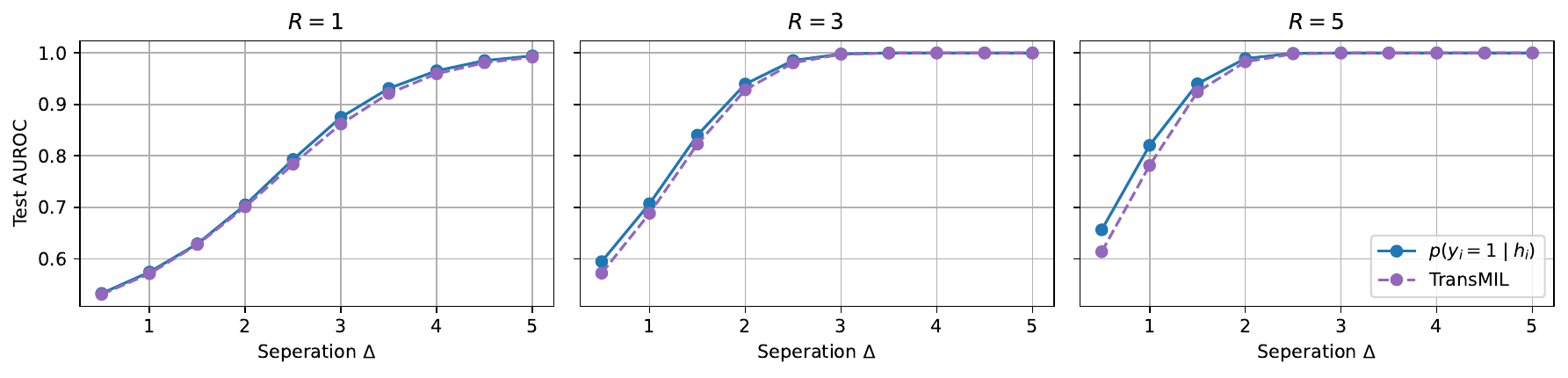}
  \caption{Handcrafted parameters (\emph{embedding-aggregation} approach).
  }
  \label{fig:handcrafted_parameters2}
\end{figure*}
\begin{figure*}[htbp!]
  \centering
  \includegraphics[width=1\linewidth]{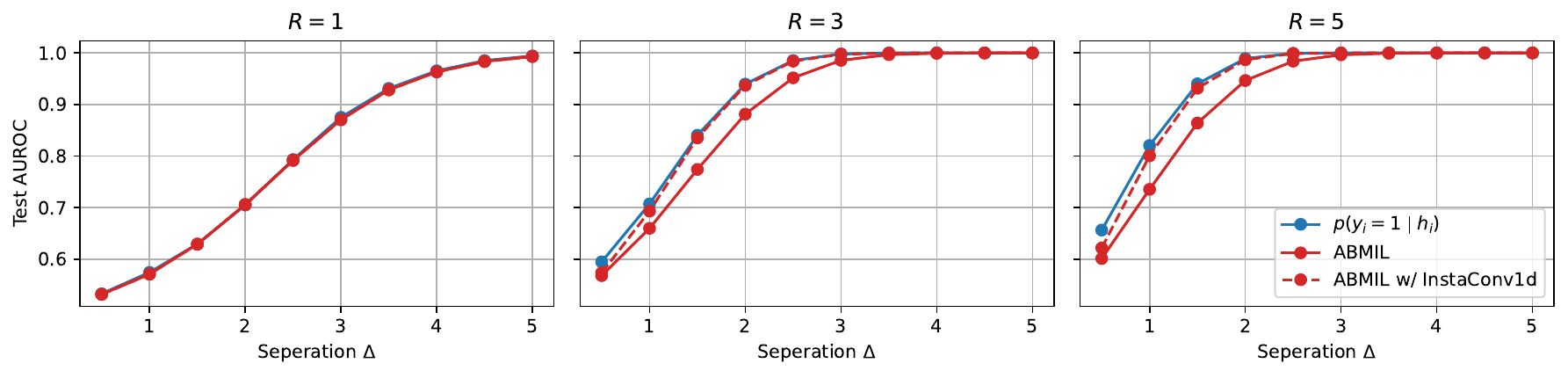}
  \caption{Handcrafted parameters (\emph{prediction-aggregation} approach).
  }
  \label{fig:handcrafted_parameters3}
\end{figure*}

\newpage
\section{Bootstrap Analysis}
\label{sec:bootstrap_analysis}

We use bootstrapping \citep{foody2009classification} to access the statistical significance of the AUROC difference between the Bayes estimator and smAP for the \emph{prediction-aggregation} approach trained with $N{=}10000$.
We report the mean and 95\% confidence interval over 500 subsamples of the test set.

\begin{figure*}[htbp!]
  \centering
  \includegraphics[width=0.5\linewidth]{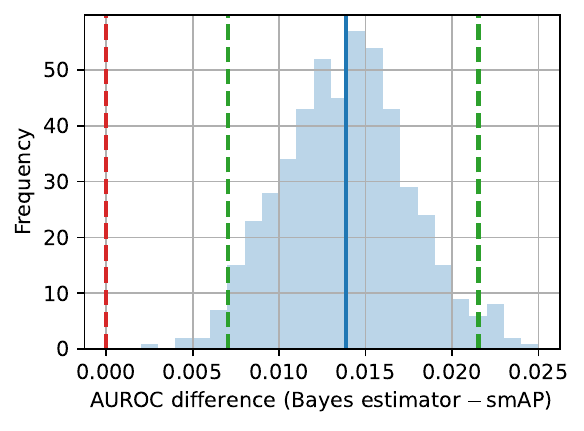}
  \caption{Bootstrap analysis comparing the AUROC difference between the Bayes estimator and smAP for the \emph{prediction-aggregation} approach trained with $N{=}10000$.
  }
  \label{fig:bootstrap_analysis}
\end{figure*}

\section{Details Needed for Derivation of Bayes Estimator}
\label{sec:distribution_over_number_of_slices}

One reviewer asked whether the generative process for the number of instances per bag $S_i$ needs to be accounted for in the Bayes estimator derivation. We can show that because this distribution is uniform and not dependent on class label, that is $p(S_i, y_i) = p(y_i)p(S_i)$, the relevant term can be factored out of the numerator and denominator in the label-given-features posterior and canceled
\begin{align}
    p(y_i=1 \mid h_i, S_i) = \frac{p(h_i \mid y_i=1, S_i) p(y_i=1) \cancel{p(S_i)}}{p(h_i \mid y_i=0, S_i) p(y_i=0) \cancel{p(S_i)} + p(h_i \mid y_i=1, S_i) p(y_i=1) \cancel{p(S_i)}}.
\end{align}

For brevity, we omitted the dependence on $S_i$ in many statements in the main paper. The complete class-conditional likelihood PDFs needed to evaluate the right-hand-side above are:
\begin{align}
    p( h_i | y_i = 0, S_i) &= \prod_{j=1}^{S_i} \prod_{k=1}^M \text{NormPDF}( h_{ijk} | \mu, \sigma^2 )
\\
    p( h_i | y_i = 1, S_i) &= \sum_{u=1}^{S_i - R + 1} \left[ \underbrace{\frac{1}{S_i {-} R {+} 1}}_{p(u | y_i=1, S_i)} \prod_{j=1}^{S_i} \prod_{k=1}^M \underbrace{\text{NormPDF}( h_{ijk} | \mu + \Delta \delta_{u,j,k}, \sigma^2 )}_{p( h_{ijk} | u, y_i=1, S_i)}
    \right]
\end{align}
where $\delta_{u,j,k}$ is 1 if $j \in [u, u+R - 1]$ and $k$ is a discriminative feature, and 0 otherwise.

\end{document}